\documentclass{article}

\usepackage{arxiv}
\usepackage[utf8]{inputenc} 
\usepackage[T1]{fontenc}    
\usepackage{url}            
\usepackage{nicefrac}       
\usepackage{microtype}      
\usepackage{doi}
\usepackage{textcomp}
\usepackage{times}
\usepackage{xcolor}
\usepackage{times}
\usepackage{epsfig}
\usepackage{graphicx}
\usepackage{amsmath}
\usepackage{amssymb}
\usepackage{amsfonts}
\usepackage{booktabs}
\usepackage{amsfonts}
\usepackage{mathrsfs}
\usepackage{tabularx}
\usepackage{mathtools}
\usepackage{float}
\usepackage{bm}
\usepackage{footnote}
\usepackage{colortbl}
\usepackage{multirow}
\usepackage{wasysym}
\usepackage{nth}
\usepackage{xspace}
\usepackage{csquotes}
\usepackage[ruled,vlined]{algorithm2e}
\usepackage[noend]{algpseudocode}

\usepackage[noadjust]{cite} 

\makeatletter
\let\NAT@parse\undefined
\makeatother
\usepackage{hyperref}

\renewcommand{\vec}[1]{\bm{#1}} 
\newcommand{\im}[1]{\bm{I}_{\text{#1}}} 
\newcommand{\oc}[1]{\bm{O}_{\text{#1}}} 
\newcommand{\msk}[1]{\bm{M}_{\text{#1}}} 
\newcommand{\Smean}[1]{\overline{\bm{S}}_{\text{#1}}} 
\newcommand{\occ}[1]{\texttt{occ}(#1)} 
\newcommand{\dcos}{d} 

\makeatletter
\DeclareRobustCommand\onedot{\futurelet\@let@token\@onedot}
\def\@onedot{\ifx\@let@token.\else.\null\fi\xspace}
\def\eg{\emph{e.g}\onedot} 
 
\def\cf{\emph{cf}\onedot}

\def\etal{\emph{et al}\onedot}

\makeatother

\makesavenoteenv{tabular}
\makesavenoteenv{table}
\makeatletter

\title{Explainable Model-Agnostic Similarity and Confidence in Face Verification}

\author{\href{https://orcid.org/0000-0002-0503-4600}{\includegraphics[scale=0.06]{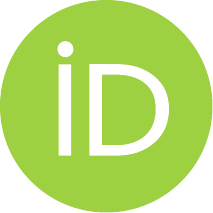}\hspace{1mm}Martin Knoche} \qquad \href{https://orcid.org/0000-0002-2963-108X}{\includegraphics[scale=0.06]{orcid.pdf}\hspace{1mm}Torben Teepe} \qquad Stefan Hörmann \qquad \href{https://orcid.org/0000-0003-1096-1596}{\includegraphics[scale=0.06]{orcid.pdf}\hspace{1mm}Gerhard Rigoll}\\
Technical University of Munich\\
Arcisstrasse 23, 80333 M\"unchen, Germany\\
	\texttt{\href{Martin.Knoche@tum.de}{Martin.Knoche@tum.de}} \\
}

\renewcommand{\headeright}{A Preprint}
\renewcommand{\undertitle}{A Preprint \\
Published in Winter Conference on Applications of Computer Vision Workshop 2023 \\ 
DOI: \url{https://doi.org/10.1109/WACVW58289.2023.00078}}
\renewcommand{\shorttitle}{Explainable Face Verification}

\hypersetup{
pdftitle={explainable face verification},
pdfsubject={},
pdfauthor={Martin~Knoche, Torben~Teepe, Stefan~H\"ormann, Gerhard~Rigoll},
pdfkeywords={face, recognition, verification, explainable, xai, artificial, intelligence, facial, visualization, confidence, score, platform},
}

\makeatletter
\providecommand*{\input@path}{}
\g@addto@macro\input@path{{./figures/}{./contents/}}
\makeatother

\usepackage[capitalize, nameinlink]{cleveref}
\crefname{section}{Sec.}{Secs.}
\Crefname{section}{Sec.}{Secs.}
\crefname{subsection}{Subsec.}{Subsecs.}
\Crefname{subsection}{Subsec.}{Subsecs.}
\Crefname{table}{Table}{Tables}
\crefname{table}{Table}{Tables}
\Crefname{figure}{Fig.}{Figs.}
\crefname{figure}{Fig.}{Figs.}
\Crefname{equation}{Equation}{Equations}
\crefname{equation}{Equation}{Equations}

\begin{document}
\maketitle

\setlength{\abovedisplayskip}{9pt}
\setlength{\belowdisplayskip}{9pt}

\begin{abstract}
	Recently, face recognition systems have demonstrated remarkable performances and thus gained a vital role in our daily life. They already surpass human face verification accountability in many scenarios. However, they lack explanations for their predictions. Compared to human operators, typical face recognition network system generate only binary decisions without further explanation and insights into those decisions. This work focuses on explanations for face recognition systems, vital for developers and operators. 
First, we introduce a confidence score for those systems based on facial feature distances between two input images and the distribution of distances across a dataset.
Secondly, we establish a novel visualization approach to obtain more meaningful predictions from a face recognition system, which maps the distance deviation based on a systematic occlusion of images. The result is blended with the original images and highlights similar and dissimilar facial regions. 
Lastly, we calculate confidence scores and explanation maps for several state-of-the-art face verification datasets and release the results on a web platform. We optimize the platform for a user-friendly interaction and hope to further improve the understanding of machine learning decisions. The source code is available on GitHub\footnote{\url{https://github.com/martlgap/x-face-verification}} , and the web platform is publicly available at \url{http://explainable-face-verification.ey.r.appspot.com}.

 \end{abstract}
 
 \section{Introduction}
Machine learning has recently demonstrated remarkable performances in multiple tasks, from image processing to natural language processing, especially with the advent of deep learning. Along with research progress, it has influenced many fields and disciplines. For example, in the medical sector or security systems, a high level of accountability and thus greater transparency and interpretability is required. However, these systems are often considered black boxes, and it is not known what happens internally. They lack an explanation. According to Phillips \etal \cite{phillips2020four}, an accompanying explanation needs to be interpretable and accurate, and models must operate within their known limits. 

\begin{figure}[t]
    \includegraphics[width={\columnwidth}]{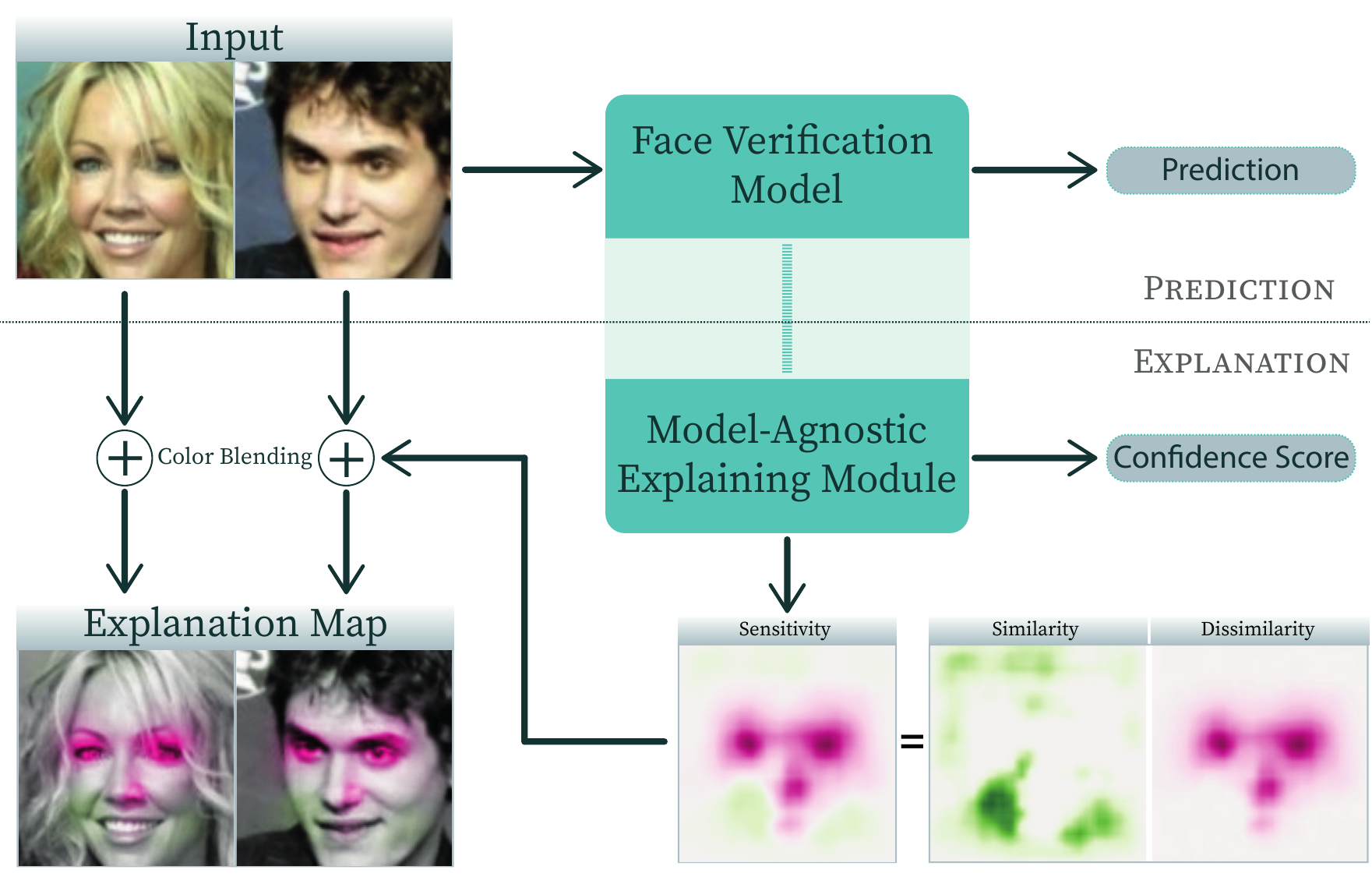}
    \caption{The proposed approach generates a similarity map and blends it with the input images into an explanation map. Besides the binary prediction of the network, we introduce a confidence score to explain the decision further.}
    \label{fig:teaser}
\end{figure}

Explainable artificial intelligence (XAI) arose from the need to understand models in various areas~\cite{thakur2022explainable, franco2022deep}. The benefits of explanations are apparent and in recent years, more and more approaches have been introduced. With ever more explainable face recognition systems, humans are getting more involved in the decision process, which is vital for many fields of applications. But explainability is not only important for the final user, but also for developers, which can benefit from a better understanding of datasets and models. The distribution and accessibility of those explanations is necessary. 

There exist model-agnostic approaches~\cite{mery2022black, mery2022true} and the famous model-specific Gradient-weighted Class Activation Mapping approaches (GradCAM) \cite{selvaraju2017grad, draelos2020hirescam, wang2020score}, which propose saliency maps, highlighting decisive facial regions. They often require access to the layers of deep learning architectures used by facial matchers, which is not always feasible in commercial systems. 
Instead, we follow the idea of Mery~\cite{mery2022black} and consider deep learning models as input-output functions, which cannot be accessed. Our model-agnostic approach focuses on the face verification problem, and we provide explanations with similarity maps, which indicate similar and dissimilar regions of the face. Instead of simply interpreting the cosine distance for the certainty of the decision, we establish a more precise confidence score calculation (see~\cref{fig:teaser}).\\
Our main contributions are summarized as follows:
\begin{itemize}
    \item We introduce a confidence score for face recognition networks.
    \item We provide three different explanation methods for face recognition.
    \item We build and release a user-friendly, interactive, and modern web platform containing the proposed confidence scores and explanation maps for several state-of-the-art datasets and models. 
\end{itemize}

 \section{Related Work}
\subsection{Explanation Maps}

One of the earliest approaches for explainable artificial intelligence (XAI) is the local interpretable model-agnostic explanations (LIME) technique introduced by Ribeiro \etal~\cite{ribeiro2016should}. In their work, they proposed a method for faithfully explaining any classifier's predictions by learning an interpretable model locally around the prediction. 

The most relevant XAI methods similar to our approach are model-agnostic algorithms: 

Firstly, Mery and Morris~\cite{mery2022black} introduced six different saliency maps that can be used to explain any face verification algorithm without manipulating the model. The key idea of their method is to define a matching score of two facial images, which changes when one image is perturbed. In addition, they experimented with XAI saliency maps based on contours. 

Secondly, in \cite{mery2022true}, Mery introduced an XAI method based on how the probability of recognition of a given image changes when it is perturbed. His algorithm removes and aggregates different parts of the image and then measures the contributions of those parts individually and in-collaboration as well. The generated saliency maps highlight the most relevant areas for the recognition process. 

Third, the work from Lin \etal~\cite{lin2021xcos} provided a learnable module that can be integrated into most face verification models. This module generates meaningful explanations with the help of a patched cosine map and an attention map. These maps represent similarities instead of saliency. 

Other model-specific XAI techniques require knowledge of the structure to observe or manipulate the outputs of hidden model layers: 
The most popular approach is the Gradient-weighted Class Activation Mapping (Grad-CAM)~\cite{selvaraju2017grad} algorithm that utilizes the gradient of the class signal with respect to the input image; 
Recently, many other XAI techniques based on GradCAM, like GradCAM++~\cite{chattopadhay2018grad}, HiResCAM~\cite{draelos2020hirescam}, AblationCAM~\cite{ramaswamy2020ablation}, ScoreCAM~\cite{wang2020score}, or XGradCAM~\cite{fu2020axiom}, have been introduced; 
Cao \etal~\cite{cao2015look} modified a network with a feedback loop to infer the activations of hidden layers according to the corresponding targets;
In \cite{li2018tell} and \cite{dabkowski2017real}, the authors trained separate models to predict saliency explanation maps;
Pruning a neural network for a given single input to keep only neurons that highly contribute to the prediction was introduced in the work of Khakzar \etal~\cite{khakzar2019improving}.

\subsection{Confidence Scores}
In~\cite{huber2022stating} Huber \etal exploited the approximation of model uncertainty through dropout and proposed an uncertainty score for the comparison of two images. Based on that, they additionally calculated a decision confidence to make the decisions for face verification more transparent without any training effort. 

In contrast, Li \etal~\cite{li2021spherical} propose a novel framework for face confidence learning in a spherical space. They extended the Mises Fisher density to it´s r-radius counterpart. 
 \section{Method}
\label{sec:method}

\subsection{Confidence Score}
\label{sec:conf_score}
Nowadays, face verification systems~\cite{deng2019arcface, knoche2022octuplet,zhong2021facetransformer,meng2021magface, liu2022anchorface, boutros2022elasticface} make predictions based on the distance between two feature vectors. Those feature vectors are typically derived from a convolutional neural network $\mathscr{N}(\cdot)$, which extracts facial features $\mathscr{N}(\im{}) = \vec{f}$ from an aligned facial image $\im{} \in {\mathbb{R}^{112\times112\times3}}$. Most approaches utilize the cosine distance metric $\dcos$ for calculating the distance between two facial feature vectors $\vec{f}_1,\vec{f}_2$ which is defined as:
\begin{equation}
    \dcos(\vec{f}_1,\vec{f}_2) = 1 - \frac{\vec{f}_1 \cdot \vec{f}_2}{\lVert \vec{f}_1 \rVert_2 \, \lVert \vec{f}_2 \rVert_2}. 
\end{equation}
From this follows that $\dcos \in [0, 2]$, whereas $\dcos$ is $0$ for identical features, $1$ for orthogonal vectors, and $2$ for opposite vectors. To classify a pair of images as genuine ($\dcos \le t$) or imposter ($\dcos > t$) one can then define a particular threshold $t$. For common face verification benchmark datasets (\eg, LFW~\cite{huang2014lfw}, CALFW~\cite{zheng2017calfw}, CPLFW~\cite{zheng2018cplfw}, SLLFW~\cite{deng2017sllfw}, XQLFW~\cite{knoche2021xqlfw}), the threshold $t$ is derived by applying 10-fold cross-validation on the test set. A certain threshold is found for each fold by maximizing the verification accuracy on the remaining folds. A prediction from a face verification with a distance $\dcos$ close to the threshold $t$ can be interpreted as uncertain. In contrast, a large distance close to $2$ or a small distance close to $0$ indicates high confidence in the model's prediction.

However, there is no clear rule on interpreting the absolute distance to the threshold $t$ in terms of prediction confidence. For instance, the FaceTransformer~\cite{zhong2021facetransformer} model in the work of Knoche \etal~\cite{knoche2022octuplet} has a threshold of $t \approx 0.2$, leading to highly imbalanced thresholds in the interval. 

In this work, we aim for a more expressive metric and introduce a confidence score (C-Score) $s$, which takes not only this imbalance into account, but also exploits information from the distribution of correct and wrong predictions of the model for each dataset. Our C-Score is calculated as follows:

\begin{figure}[t]
    \includegraphics[width={\columnwidth}]{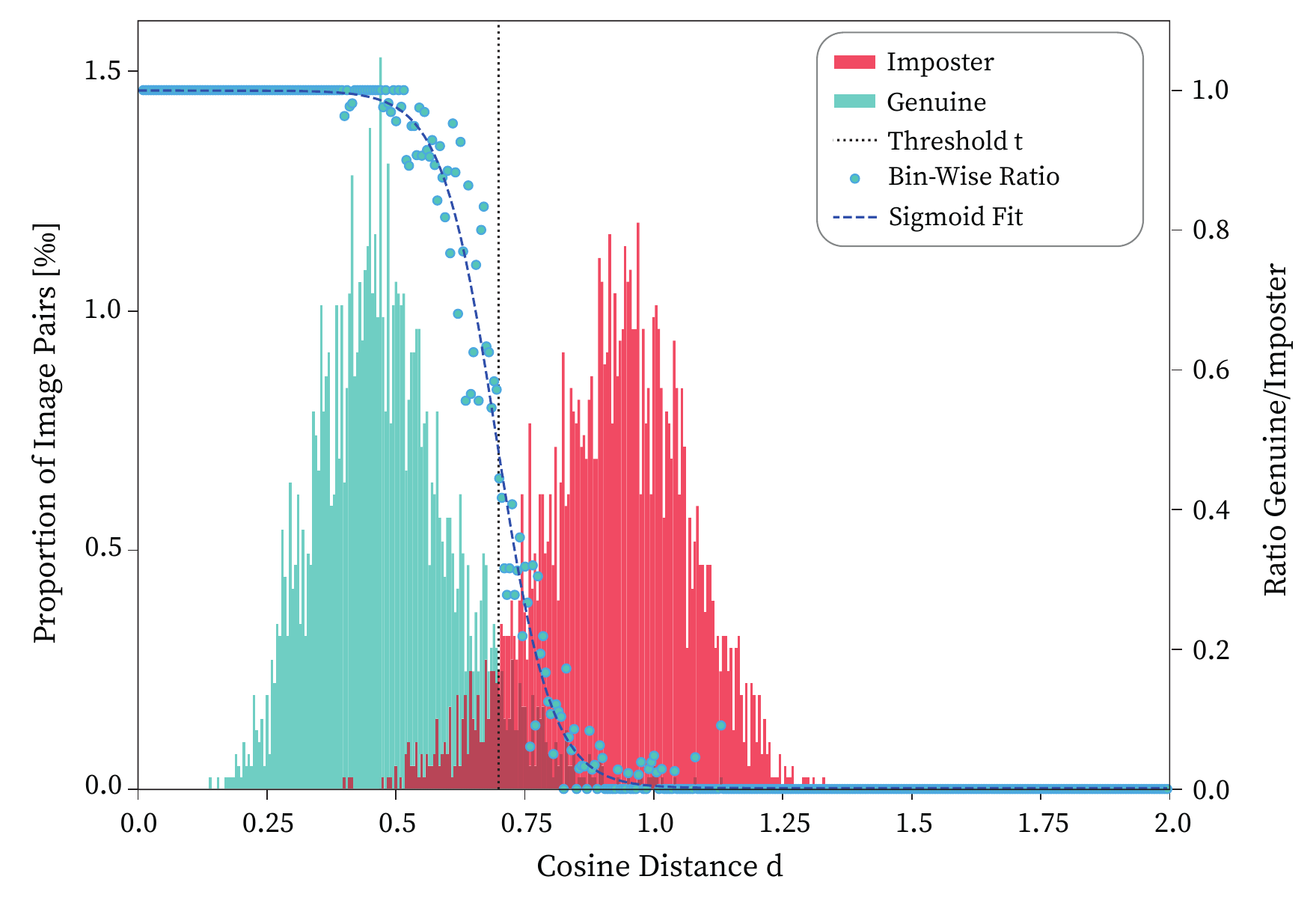}
    \caption{Histogram of cosine distances for the first fold of the LFW~\cite{huang2014lfw} dataset and the bin-wise ratio between genuine and imposter distance counts. The distances are derived from an ArcFace~\cite{deng2019arcface} model fine-tuned with OctupletLoss~\cite{knoche2022octuplet}.}
    \label{fig:hist}
\end{figure}

Given the cosine distance distribution derived from an arbitrary face verification model for genuine and imposter image pairs, we compute the histogram (\cf~\cref{fig:hist}) with $400$ bins. The bin-wise ratio of the number of genuine examples to imposter examples follows an s-shaped distribution starting from $1$ for $\dcos < t$ and ending at $0$ for $\dcos > t$. The closer $\dcos$ gets to the threshold $t$, the more uncertain the prediction is due to more misclassifications in that range. We interpret the left part of the distribution ($\dcos < t$) as the probability that a given genuine prediction is correct and the right part of the distribution ($\dcos > t$) as the improbability that a given imposter prediction is correct. Then, we fit a logistic sigmoid curve $c(\dcos)$ depending on the parameters $L, d_0, k, b$, defined as,
\begin{equation}
c = \frac{L}{(1 + e^{-k \cdot (\dcos-d_0)})} + b
\end{equation}
to the distribution of ratio values using the dogbox~\cite{voglis2004rectangular} algorithm. This enables a continuous mapping of arbitrary distance values. Because the fitted sigmoid curve $c$ is an approximation, we clip the $c$ to a range $[0,1]$. Finally, to get a more intuitive score, we invert the improbability values and define our C-Score $C$ as: 
\begin{equation}
    C = \begin{cases}
        c(d) & \forall \quad \dcos \le t \\
        1-c(d) & \forall \quad \dcos > t
    \end{cases}
\end{equation}
As a result, we obtain our introduced C-Score $C$ in the range of $[0.5, 1]$ for either genuine or imposter predictions and can interpret it as a probability for correctness. Notice that the calculation of $C$ is done fold-wise, resulting in altering parameters for each fold of the dataset. 

Finally, with C-Score $C$, we establish an additional value to the binary output prediction of a face verification system and thus make the prediction more meaningful. It is important to note that for the C-Score $C$, we gathered ground truth information of the dataset; hence, for the application of the model to field data, the parameters for the C-Score function need to be derived from a validation dataset.

\subsection{Model-Agnostic Explanation Maps}
\label{sec:explanation_maps}
The core principle of our model-agnostic explanation approach is visualizing the deviation between a non-occluded and an occluded image. If the feature distance between those two images is decreasing, we interpret the occluded area as dissimilar and vice versa similar for a greater distance. With systematic image occluding, we can then measure the influence on the cosine distance of every part of the image and hence, visualize this in a 2-D map. 

In the following, the procedure is explained in more detail: 1) First, we apply our proposed \cref{algo:img_occ}, which can be formulated as 
\begin{equation}
    \begin{aligned}
        \occ{\im{}} \mapsto \quad \mathcal{O} & := \big\{\oc{1},\oc{2},\dots,\oc{N}\big\}, \\ \mathcal{M} & := \big\{\msk{1},\msk{2},\dots,\msk{N} \big\}
    \end{aligned}
\end{equation}
with $\oc{} \in \mathbb{R}^ {112\times112\times3}$, $\msk{} \in \mathbb{R}^{112\times112}$, and $N = \lfloor (112-p) / s \rfloor^2$, dependent on the patch size $p$ and a stride $s$. Note that our masks $\msk{}$ are sparsely populated; only the occluded areas contain the values $1$. After utilizing $\occ{\cdot}$ on our input image 2-tuple ($\im{1}, \im{2}$), we retrieve a 2-tuple of occluded image sets ($\mathcal{O}_1, \mathcal{O}_2$) and a 2-tuple of mask sets ($\mathcal{M}_1, \mathcal{M}_2$). 

\begin{algorithm}[t]
    \SetAlgoLined
    \KwIn{image \textbf{\textit{I}}}
    $s \longleftarrow \text{stride}$\\
    $p \longleftarrow \text{size of patch}$\\
    Start at top left corner of \textbf{\textit{I}} \\
    \While(move right \textit{s} pixels){within \textbf{\textit{I}}}{
        \While(move down \textit{s} pixels){within \textbf{\textit{I}}}{
            $\msk{} \longleftarrow \text{draw a patch with size } p \text{ at } loc$\\
            $\oc{} \longleftarrow \text{occlude } \textbf{\textit{I}} \text{ with patch of size }p \text{ at } loc$\\
        }
    }
    \KwOut{occluded images $\mathcal{O}$, masks $\mathcal{M}$}
    \caption{Systematic Image Occluding $\occ{\cdot}$\label{algo:img_occ}}
\end{algorithm}

In the next step, we extract the facial features $\vec{f} \in \mathbb{R}^{512}$ with a face verification network $\mathscr{N}(\cdot)$ for every single occluded image $\oc{}$ in the 2-tuple ($\mathcal{O}_1, \mathcal{O}_2$): 
\begin{equation}
    \mathcal{F} := \big\{\vec{f}_1,\vec{f}_2,\dots,\vec{f}_N : \vec{f} = \mathscr{N}(\oc{})\big\}, 
\end{equation}
and consequently, generate a 2-tuple of feature vector sets $(\mathcal{F}_1, \mathcal{F}_2)$. Then, we calculate the cosine distance $\dcos(\cdot\,,\cdot)$ between all features in $\mathcal{F}_{1}$ and $\mathcal{F}_{2}$. To select the 2-tuple of pair-wise distances sets ($\mathcal{D}_1, \mathcal{D}_2$), we employ the following three methods:

\BlankLine
\textbf{Method 1} selects the cosine distances $\mathcal{D}_1$ and $\mathcal{D}_2$ according to
\begin{equation}
    \begin{aligned}
    &\mathcal{D}_1 := \Bigg\{\sum^{N}_{j=1}\frac{\dcos(\mathcal{F}^{(i)}_1,\mathcal{F}^{(j)}_2)}{N}: \forall\,i \in [1,2,\dots,N] \Bigg\} \\
    &\mathcal{D}_2 := \Bigg\{\sum^{N}_{i=1}\frac{\dcos(\mathcal{F}^{(i)}_1,\mathcal{F}^{(j)}_2)}{N}: \forall\,j \in [1,2,\dots,N] \Bigg\}.
    \end{aligned}
    \label{eq:method1}
\end{equation}
With this selection, we extract the averaged influence of all occluded (at any location) image for one of the input images compared with the occluded (at a particular location) image of the other input image. 

\BlankLine
\textbf{Method 2} selects the cosine distances $\mathcal{D}_1$ and $\mathcal{D}_2$ according to
\begin{equation}
    \begin{aligned}
    &\mathcal{D}_1 := \big\{\dcos(\mathcal{F}^{(i)}_1,\mathscr{N}(\im{2})): \forall\,i \in [1,2,\dots,N] \big\} \\
    &\mathcal{D}_2 := \big\{\dcos(\mathscr{N}(\im{1}),\mathcal{F}^{(i)}_2): \forall\,i \in [1,2,\dots,N] \big\}.
    \end{aligned}
    \label{eq:method2}
\end{equation}
This selection aims to measure the influence of one of the input images compared with the occluded (at any location) image of the other input image. 

\BlankLine
\textbf{Method 3} selects the cosine distances $\mathcal{D}_1$ and $\mathcal{D}_2$ according to
\begin{equation}
    \mathcal{D}_1 = \mathcal{D}_2 := \big\{\dcos(\mathcal{F}^{(i)}_1,\mathcal{F}^{(i)}_2) : \forall\,i \in [1,2,\dots,N]\big\}.
    \label{eq:method3}
\end{equation}
Here, we measure the distances between the co-located occludings of both input images. 

Independent of the above-described methods, we obtain a 2-tuple of distance sets ($\mathcal{D}_1, \mathcal{D}_2$), which is then compared with the original distance $\dcos_{orig} = \dcos(\im{1},\im{2})$ of both non-occluded input images. The difference in the distance in $\mathcal{D}_1$ or $\mathcal{D}_2$ compared with $\dcos_{orig}$ is the weight for it's corresponding occlusion mask in $\msk{}$. After building the mean across all weighted masks, we generate similarity maps $\vec{S}$:
\begin{equation}
    \vec{S} = \sum^{N}_{i=1}\frac{(d_i - \dcos_{orig}) \cdot \msk{i}}{N}
    \label{eq:dev_map}
\end{equation}
with $d_i \in \mathcal{D}$ and $\msk{i} \in \mathcal{M}$. This allows visualizing the deviation caused by an occlusion at a particular location. The procedure described above is performed separately for every particular occlusion patch size $p \in \{7, 14, 28\}$ and a stride $s = 5$. The stride $s$ reduces the number of images inferred by a factor of $s^2$. Consequently, we get three similarity maps $\vec{S}$ each for both input images $\im{1}$ and $\im{2}$. Finally, we calculate the weighted average of the similarity maps based on the size of the patch area: 
\begin{equation}
    \Smean{} = \sum^{|p|}_{i=1}\frac{\vec{S}_i}{p^2_i \cdot |p|}
    \label{eq:mean_patch}
\end{equation} 

The occurring raster artifacts, caused by using a stride instead of shifting the occlusion patch pixel by pixel, are compensated by applying a Gaussian-Blur to the mean similarity maps $\Smean{}$ with an $s\times s$ kernel and $\sigma = s$, followed by normalization to the range $[-1, 1]$. 

Ultimately, we generate a 2-tuple of X-Maps for a 2-tuple of input images ($\im{1}, \im{2}$) via color blending $\texttt{blend}(\im{},\Smean{})$ (see \cref{algo:col_blend}) with the corresponding $\Smean{1}$ and $\Smean{2}$. 
\begin{algorithm}[t]
    \KwIn{image \textbf{\textit{I}}, similarity map \textit{map}}
    $l \longleftarrow \text{get luminance from: }\text{RGBtoHLS}(img)$\\
    $h \longleftarrow \text{get hue from: } \text{RGBtoHLS}(map)$\\
    $s \longleftarrow \text{get saturation from: }\text{RGBtoHSV}(map)$\\
    $\textbf{\textit{I}}_b \longleftarrow \text{HLStoRGB}(h,l,s)$\\
    \KwOut{blended image $\textbf{\textit{I}}_b$}
    \caption{Color Blending $\texttt{blend}(\cdot\,,\cdot)$\label{algo:col_blend}}
\end{algorithm}

The proposed approach generates an image-specific X-Map for both images of an arbitrary image pair. It highlights the similar and dissimilar regions of an image in terms of their identity features extracted from a face verification model.

 \section{Results}
\label{sec:results}

\subsection{Qualitative Results}
\label{sec:qualitative_results}
This section provides X-Maps for a small selection of image pairs from the LFW~\cite{huang2014lfw} dataset. With the release of our proposed \textit{eXplainable Face Verification} platform, its very easy to browse through all the generated X-Maps for several models. 

\begin{figure}[h!]
    \centering
    \includegraphics[width={0.8\columnwidth}]{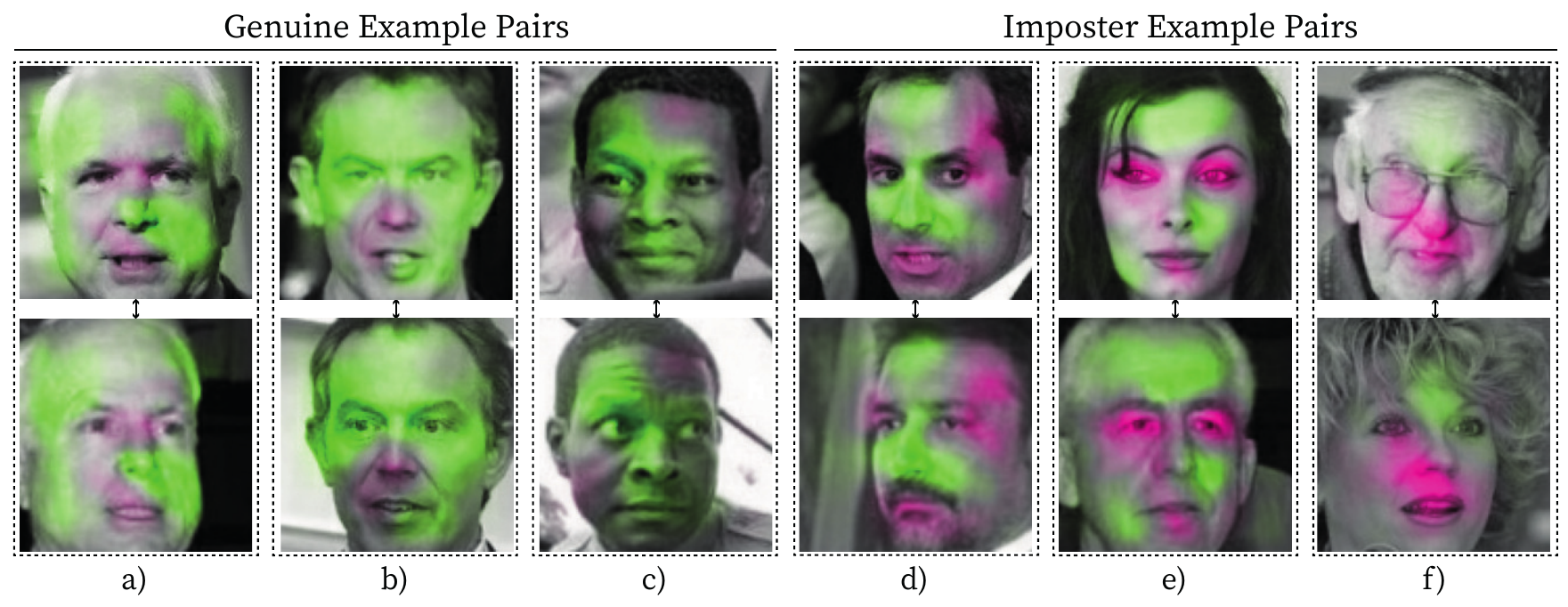}
    \caption{X-Maps for three genuine and three imposter example pairs from the LFW~\cite{huang2014lfw} dataset. Green colors indicate similar facial regions and red highlights dissimilar ones. All X-Maps are generated utilizing a FaceTransformer~\cite{zhong2021facetransformer} model fine-tuned with OcutpletLoss~\cite{knoche2022octuplet}.}
    \label{fig:overlays}
\end{figure}

The X-Maps of the genuine pairs in \cref{fig:overlays} are dominated by green-colored facial regions, for indicating similarity. In example a), the X-Map reveals that the eyes and mouth of the subject seem to not play an essential role in the model's decision. The cosine distance will get even smaller for occlusions on those parts of the face. In b), the nose is the only facial part, which is less critical for the model's prediction. In the genuine example pair c), eyes, nose, and mouth are highlighted, and push the distance closer to zero. 

Not surprisingly, the X-Maps of the imposter pairs indicate more dissimilar facial regions than the genuine pairs' X-Maps. Whereas the nose of the subjects in pair f) is marked very dissimilar, that is the case for the eyes of pair e). Interestingly, the nose of pair d) is specified as a very similar facial region. 

All three X-Maps indicate that the forehead is rather similar compared to the more distinctive facial parts such as the eyes, nose, and mouth. 

\begin{figure}[h!]
    \centering
    \includegraphics[width={0.8\columnwidth}]{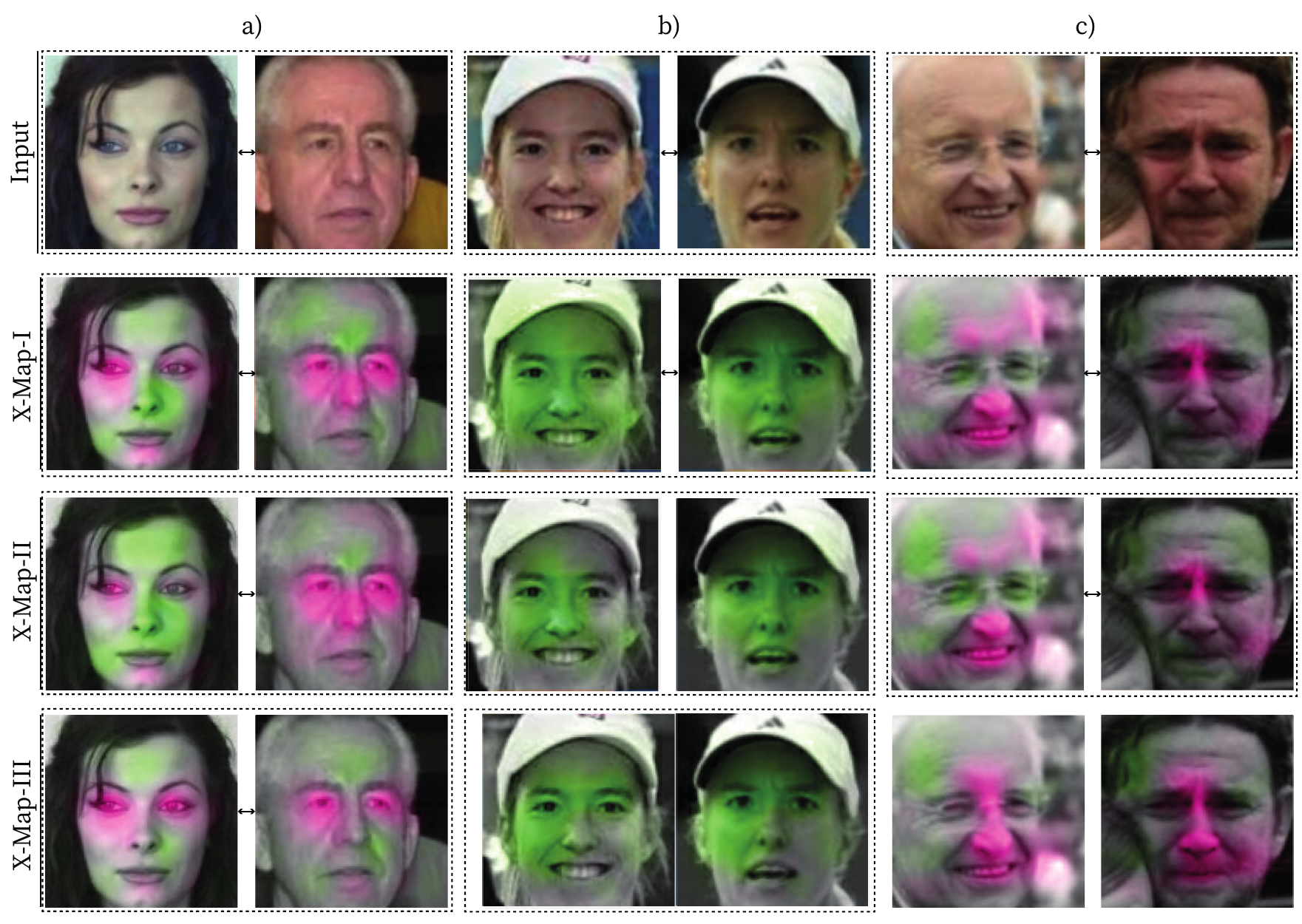}
    \caption{Comparison of our three proposed explanation maps algorithms for three example image pairs of the LFW~\cite{huang2014lfw} dataset. Green colors indicate similar facial regions and red highlights dissimilar ones. All X-Maps are generated utilizing a FaceTransformer~\cite{zhong2021facetransformer} model fine-tuned with OcutpletLoss~\cite{knoche2022octuplet}.}
    \label{fig:method_comp}
\end{figure}

\subsection{Comparison of Explanation Maps}

In this section, we compare our three proposed methods (\cf~\cref{sec:explanation_maps}) of generating the X-Maps. Whereas the X-Map-I (\cf~\cref{eq:method1}) technique considers both images occluded, the X-Map-II (\cf~\cref{eq:method2}) technique compares a non-occluded image with an occluded image. The main difference between those two methods and X-Maps-III (\cf~\cref{eq:method3}) are the co-located occlusions in both images, resulting in identical X-Maps for both images. Therefore, X-Map-III method is most expressive for normalized and frontal facial images with co-located facial parts.

In \cref{fig:method_comp}, we present the different X-Maps for three example image pairs. 
The X-Map-I and X-Map-II methods indicate different parts of the face as similar or dissimilar, which makes it difficult for a human to interpret the explanation. Nevertheless, this enables better explainability for misaligned, varying pose, or occluded images. 
The bottom row (Method-III) reveals that the eyes in example pair a) most strongly impact the prediction into the imposter direction. In c), the same holds for the nose and mouth region. Compared to a) where the eyes are clearly visible and of good quality, in c), they are of very bad quality and hence, are not considered playing an essential role in the verification prediction from our algorithm. In all images, the forehead region is highlighted rather as similar, which is obviously due to the lack of information. 

\subsection{Experiments with Cut-and-Paste Patches}
\begin{figure}[t]
    \centering
    \includegraphics[width={0.8\columnwidth}]{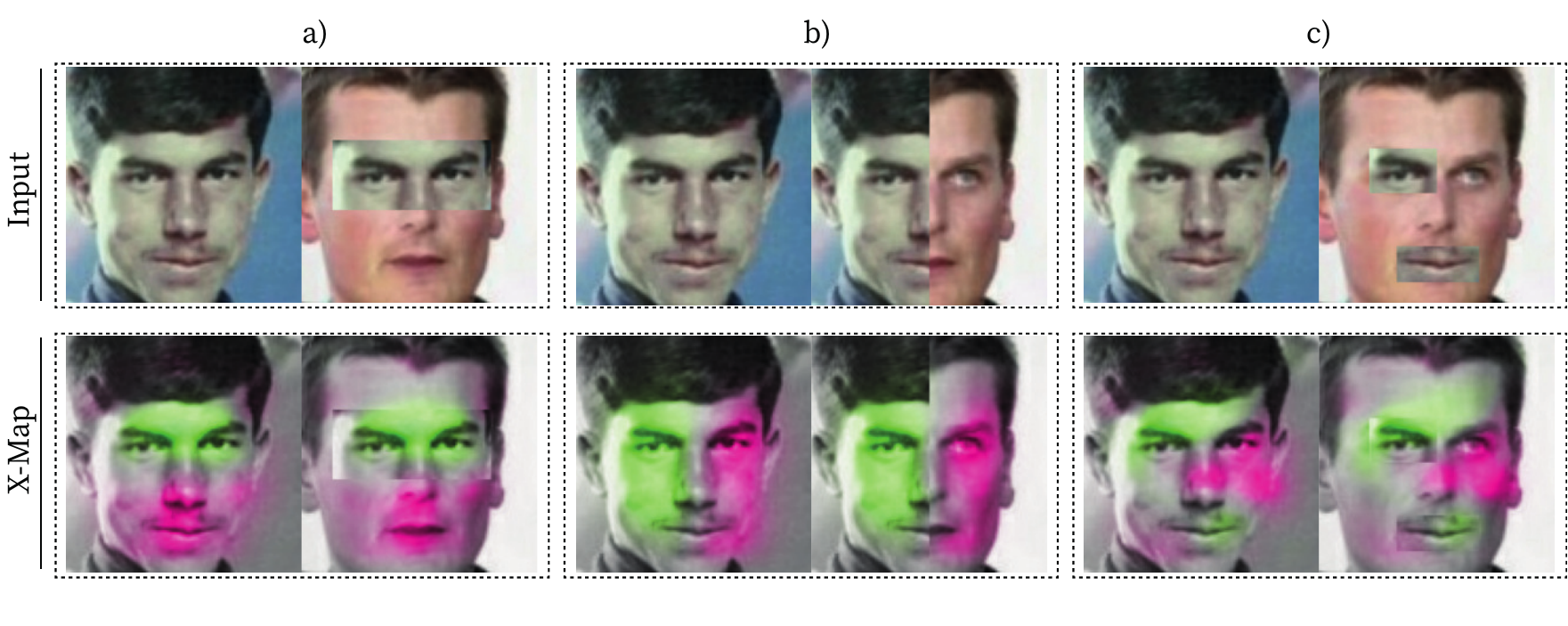}
    \caption{X-Maps (method-III) for three example image pairs of the LFW~\cite{huang2014lfw} dataset with modified facial regions. Green colors indicate similar facial regions and red highlights dissimilar ones. All X-Maps are generated utilizing a FaceTransformer~\cite{zhong2021facetransformer} model fine-tuned with OcutpletLoss~\cite{knoche2022octuplet}.}
    \label{fig:inpaintings}
\end{figure}

Additionally, we conduct experiments with modified image pairs. This experiment investigates whether the replacement of particular facial regions in one image with a copy of the co-located region of the other image is successfully detected by our algorithm and described with high similarity in the X-Maps. \cref{fig:inpaintings} depicts three facial replacements such as the eye region a), half side of the face b), or one eye and mouth c). In all three examples, our proposed algorithm highlights the copied facial regions as similar and the remaining facial area as dissimilar. In example c), this effect is most weakly pronounced. 

\subsection{Sensitivity Studies}
\label{sec:ablation}
We conduct multiple sensitivity studies to determine the influence of the size, edge quality, coloring, and shape of the patches used in our systematic image occlusion algorithm~\cref{algo:img_occ}.

\begin{figure}[h!]
    \centering
    \includegraphics[width={0.8\columnwidth}]{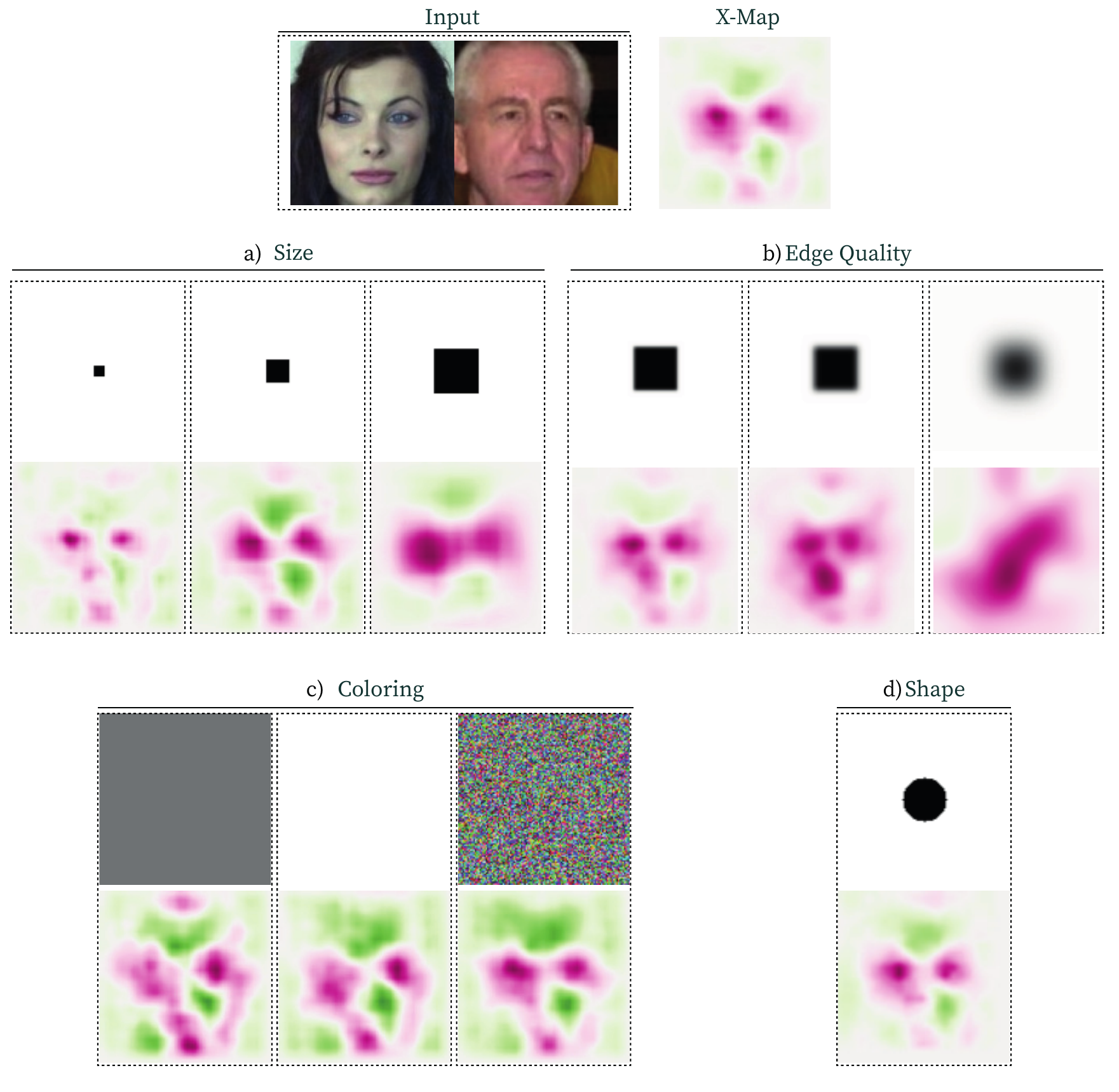}
    \caption{Similarity maps (method-III) for an example image pair of the LFW~\cite{huang2014lfw} dataset, generated with a different patch size, edge quality, coloring, and shape. Green colors indicate similar facial regions and red highlights dissimilar ones. All similarity maps are generated utilizing a FaceTransformer~\cite{zhong2021facetransformer} model fine-tuned with OcutpletLoss~\cite{knoche2022octuplet}.}
    \label{fig:ablation}
\end{figure}

First, we use three different patch sizes ($7\times7, 14\times14, 28\times28$ pixels) for our systematic image occlusion algorithm~\cref{algo:img_occ} to visualize the effect of the occluding patches. As depicted in \cref{fig:ablation} a), it strongly affects the resulting similarity maps. The smallest patch generates a more fine-grained similarity map and highlights small areas, which are not visible in the largest patch. In order to obtain one generalized X-Map (top right of \cref{fig:ablation}), with information from different levels of granularity, we merged three maps (all patches black colored, rectangular shaped and not Gaussian blurred) and weighted them based on the area of the patches (\cf~\cref{eq:mean_patch}). 

Second, we analyze three different edge qualities in the patches, in the form of different levels of Gaussian blurring. For both the kernel size and sigma, we use the values $\{7, 14, 56\}$. As seen in b), this affects the similarity maps regarding visual granularity. 

Third, we vary the coloring of the patches and indicated in c) that it affects the similarity maps. The weakest similarity map is generated for black occlusions. The difference for gray, white, and noisy occlusion is only marginal. 

Lastly, we investigate the effect of the shape of the patches. In d), we depict the X-Maps for rectangular and round shape. There are only minor differences visible in the similarity maps. We conclude that the shape of the patch has the most minor effect on our proposed X-Maps. 

In summary, our sensitivity study reveals that the X-Maps content depends on the patch characteristics, and they should be adjusted carefully to the purpose and kind of data.

 \section{Web Platform}
In this section, we will briefly describe our \textit{eXplainable Face Verification} platform for presenting all the qualitative results of our approach and also help the community familiarizing with several test datasets and different model behaviors. This platform supports the visual understanding of the proposed algorithms. We also want our results to be easily accessible and publicly available.

\begin{figure}[h!]
    \centering
    \includegraphics[width={0.73\columnwidth}]{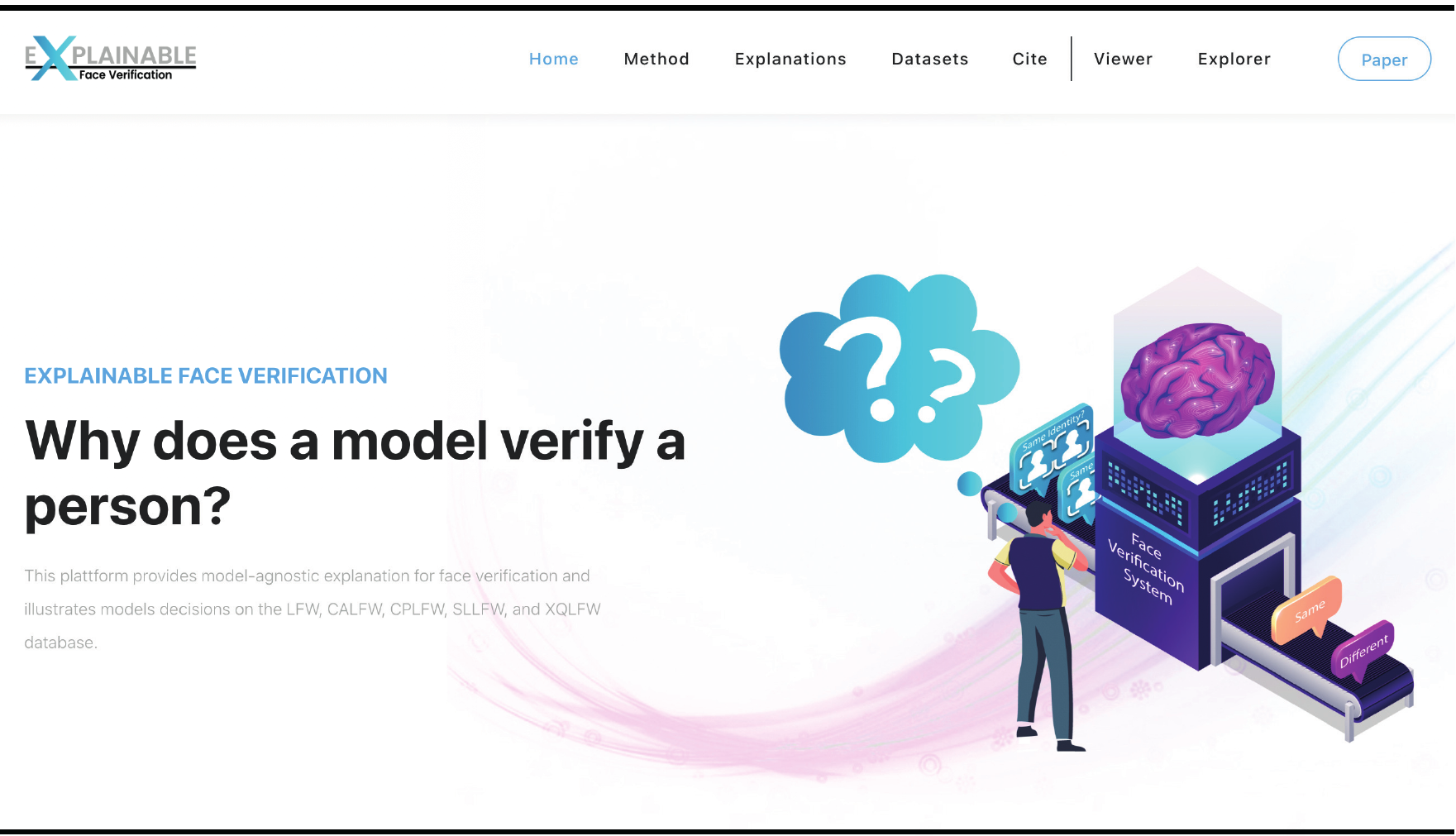}
    \caption{A screenshot of the landing page of our \textit{eXplainable Face Verification} platform}
    \label{fig:landing}
\end{figure}

\begin{figure}[h!]
    \centering
    \includegraphics[width={0.73\columnwidth}]{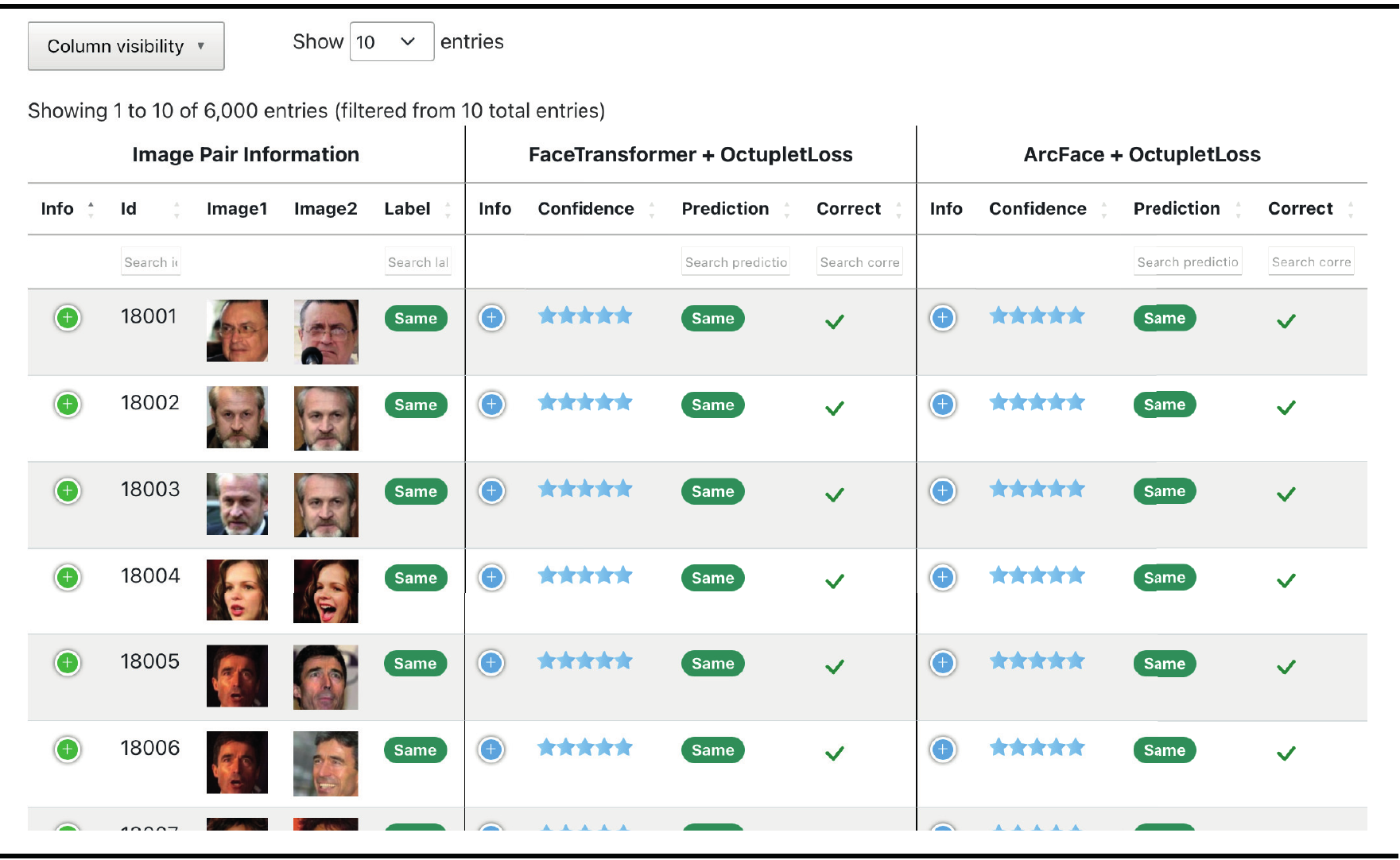}
    \caption{Screenshot of the \enquote{explorer} module of our proposed platform. It shows the interactive data table.}
    \label{fig:explorer}
\end{figure}

\begin{figure}[h!]
    \centering
    \includegraphics[width={0.73\columnwidth}]{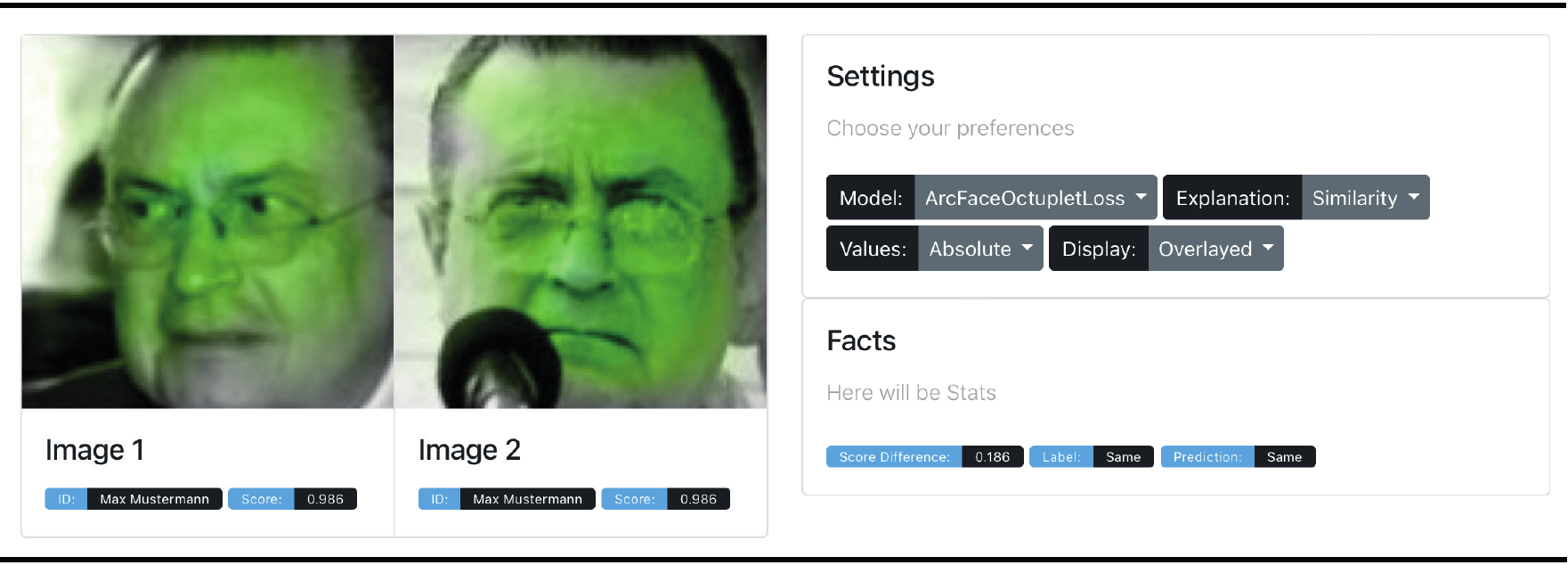}
    \caption{Screenshot of the \enquote{viewer} module of our proposed platform. It shows interactive X-Maps for an example image pair and corresponding metadata.}
    \label{fig:viewer}
\end{figure} 

The platform runs a flask~\cite{grinberg2018flask} framework, which is connected to a database containing all the datasets and the metadata. The backend does all the sorting, filtering, and accessing to improve the user experience. The platform can be divided into three parts: 

First, the landing page (see \cref{fig:landing}) explains our method and gives an overview of the accessible data and a preview of the \enquote{viewer} functionality. 

Second, the "explorer" page (see~\cref{fig:explorer}) contains an interactive table where users can filter and sort the data. The image pairs and corresponding metadata, such as file path, label, identity, and image quality score for LFW~\cite{huang2014lfw}, CALFW~\cite{zheng2017calfw}, CPLFW~\cite{zheng2018cplfw}, SLLFW~\cite{deng2017sllfw}, and XQLFW~\cite{knoche2021xqlfw}, are stored in the table. Moreover, we added the results (\eg, prediction, distance, threshold, confidence) from several face recognition models to the table.

Third, the purpose of the \enquote{viewer} page (see~\cref{fig:viewer}) is to present our generated X-Maps in an interactive, adjustable way. The user can select different models, methods, and maps for each pair of images in the datasets. 

The limitations of the \textit{eXplainable Face Verification} platform can be summarized as follows: 1) The datasets are limited to LFW and its derivatives. 2) We only applied our approach to face verification datasets. 3) The platform presents results for a small portion of existing face recognition networks. 


 \section{Conclusion and Future Work}
This work conducts further research on explainable face verification and proposes a novel strategy to generate three different explanation maps and a confidence score for a face verification model's prediction. 

With our \textit{eXplanable face verification} platform, we contribute a tool to further investigate the behavior of state-of-the-art face recognition networks and demonstrate the interpretability and accuracy of our approach.  

However, our proposed X-Map algorithm can only highlight highly locally appearing similarities. Hence, our method cannot reveal more global similarities, such as skin color or the shape of the face.

Although our work focuses explicitly on faces, the approach is not limited to the faces domain and can potentially be applied to other binary decision problems. 

In the future, we want to use the C-Scores and X-Maps for a joint application of human and machine face verification. We are planning to investigate whether a machine face verification algorithm can successfully, with the help of humans, solve the edge cases in face verification. We want to achieve this by filtering out the problematic cases based on the C-Score and using the X-Maps to support humans in their decision.

{\small
\bibliographystyle{IEEEbib}
\bibliography{ms}
}

\end{document}